\documentclass{article}

\usepackage{arxiv}
\usepackage[utf8]{inputenc}
\usepackage[T1]{fontenc}
\usepackage[hidelinks]{hyperref}
\usepackage{url}
\usepackage{booktabs}
\usepackage{amsmath,amsfonts}
\usepackage{microtype}
\usepackage{graphicx}
\usepackage[round,authoryear]{natbib}
\usepackage{tabularx}
\usepackage{array}
\usepackage{caption}
\usepackage{float}
\usepackage{tikz}
\usetikzlibrary{arrows.meta,positioning}
\usepackage{enumitem}
\setlist{nosep,leftmargin=1.4em}
\DeclareRobustCommand{\f}[1]{\texttt{\detokenize{#1}}}
\newcolumntype{L}[1]{>{\raggedright\arraybackslash}p{#1}}
\newcolumntype{Y}{>{\raggedright\arraybackslash}X}
\renewcommand{\headeright}{}
\renewcommand{\undertitle}{Preprint}
\renewcommand{\shorttitle}{PlainPairs: prescriber and patient texts of U.S. medicine labels}

\title{PlainPairs: A Corpus Pairing the Prescriber and Patient Texts of U.S.\ Prescription Medicine Labels}

\author{Obada Kraishan\\
  College of Media and Communication\\
  Texas Tech University\\
    Lubbock, TX, USA\\
  \texttt{omareikr@ttu.edu}\\
  ORCID: \href{https://orcid.org/0009-0007-7180-8620}{0009-0007-7180-8620}
}
\date{November 2026}

\begin{document}
\maketitle

\begin{abstract}
A prescription medicine in the United States comes with labeling written for prescribers, and
many medicines also carry a text written for patients: a Medication Guide or a patient package
insert. The two texts describe the same medicine for two readers, yet public resources rarely
hold them side by side. I present PlainPairs, an open corpus built from the openFDA drug label
files. Version 0.1.0 holds 35,177 prescription labels that carry a patient text, 21,416
distinct patient texts, 250,829 patient-text segments matched to prescriber sections through
their regulatory headings, reading-level measures on both sides, and rule-based counts of
plain-language wording. A second source, the National Cancer Institute's cancer information
summaries, adds 756 summaries in health-professional and patient versions, in English and
Spanish. Measured on the structure of the label, the median Medication Guide reads at grade 8.7
by Flesch-Kincaid and at grade 11.1 by SMOG. The same guides measured on the flattened text
that the bulk files serve read about two grades higher, a difference that comes from lost list
and heading boundaries. Patient texts address the reader in 41 of every 100 sentences, against
0.2 in prescriber sections. A blind hand check of 100 pairs supports the heading-based
alignment for seven of ten heading types. The corpus, its code and its documents are public,
and the release keeps texts that show a company copyright mark out of the text tables.
\end{abstract}


% ---------------------------------------------------------------------------------------
\section{Introduction}
\label{sec:intro}

People who take a prescription medicine in the United States may receive two official texts
about it. One is the prescribing information, written for clinicians. The other, for many
medicines, is a Medication Guide or a patient package insert, written for the person who takes
the medicine. Both are part of the same approved label, and both are public. They are the
closest thing the health system has to a matched pair of texts on one topic for two audiences.

That pair matters to two research communities. Health communication and health literacy
researchers have long asked whether patient medicine information can be read by the people it
is written for. Low health literacy is associated with worse health outcomes
\citep{berkman2011}, patients often misread even short instructions on medicine containers
\citep{davis2006}, and written medicine information has well-known limits as a way to inform
patients \citep{shrank2007}. Studies of Medication Guides have repeatedly found them above the
recommended reading level \citep{wolf2006,wolf2012,jones2021}. Researchers in natural language
processing, in turn, need pairs of expert and lay texts to train and test systems that rewrite
medical language for general readers \citep{althanyyan2021}, and the quality of such systems
depends on the pairs they learn from \citep{xu2015}.

Both communities have worked with small or indirect data. The readability studies cover 27 to
185 documents each, are scored once, and release no data. The simplification corpora draw
their plain side from research summaries or from text written for the corpus, not from official
patient information (Section~\ref{sec:related}). The full set of labels is public through
openFDA \citep{kasshout2016}, but the bulk files serve each label as one record with dozens of
fields, with no link between a patient text and the prescriber sections that cover the same
ground, no measures, and text that has lost its lists and headings.

This paper describes PlainPairs, a corpus that fills in those missing parts. It is built from
every label in one openFDA export, keeps the prescription labels that carry a Medication Guide
or a patient package insert, and adds four things to them:
\begin{enumerate}
  \item \textbf{Pairing and alignment.} Each patient text is cut at its standard headings, and
  each segment is matched to the prescriber sections that its heading points to, with a
  similarity score for every candidate section.
  \item \textbf{Measures on both sides.} Every text and section carries word and sentence
  counts, Flesch-Kincaid and SMOG grades, and counts of wording features such as direct
  address, commands, percentages and doses.
  \item \textbf{The label's structure.} For labels that DailyMed serves in the same version,
  paragraphs, list items and headings are read from the label's XML, so that reading levels
  are measured on the text as it is laid out and not on a flattened string.
  \item \textbf{A second pair of the same kind.} Text from the National Cancer Institute's
  cancer information summaries, in health-professional and patient versions, in English and
  Spanish.
\end{enumerate}

Three descriptive results come out of the first release. Medication Guides read at a median
grade of 8.7 by Flesch-Kincaid and 11.1 by SMOG when measured on structure, so whether they
meet a grade-8 target depends on the formula. Measuring the same guides on the flattened text
of the bulk files adds about two grades, which matters for any study that computes readability
from those files. And the patient and prescriber texts differ in the expected direction on
every measure: within the same label, the guide reads lower than the Warnings and Precautions
section in 99.9\% of 8,045 pairs.

The corpus is a record of wording. It says nothing about what a medicine does, whether it is
safe or who should take it, and it must not be used to make decisions about care. The rest of
the paper covers related work (Section~\ref{sec:related}), sources and scope
(Section~\ref{sec:sources}), how the corpus is built (Section~\ref{sec:method}), its files
(Section~\ref{sec:records}), the checks made on it (Section~\ref{sec:validation}), what the
first release shows (Section~\ref{sec:findings}), how to use it (Section~\ref{sec:usage}), and
its limits (Section~\ref{sec:limits}).

% ---------------------------------------------------------------------------------------
\section{Background and related work}
\label{sec:related}

This section describes the two patient texts the corpus is built on, the studies that measured
them, and the corpora that pair expert and lay medical text.

\subsection{Patient texts in U.S.\ prescription labeling}

Two kinds of patient text are part of approved prescription labeling. A Medication Guide is
required for medicines that the Food and Drug Administration judges to pose a serious and significant public health concern.
The regulation asks for nontechnical, understandable language without naming a reading grade,
and it sets headings in a fixed order, such as ``What is the most important information I
should know about [the medicine]?'', of which a guide carries those relevant to its product
\citep{cfr208}. A patient package insert is an older and less uniform form, long
required for a few classes of medicines and supplied voluntarily for others. Many labels also
carry Instructions for Use for products such as injectors and inhalers. In 2023 the agency
proposed to replace these forms with a single one-page document, Patient Medication
Information, phased in over five years \citep{pmi2023}. If that rule is made final, a corpus
that follows the labels over time would record the change from the old texts to the new.

The prescribing information is written for clinicians. In the format used since 2006 it has
numbered sections such as Indications and Usage, Dosage and Administration, Contraindications,
Warnings and Precautions, and Adverse Reactions; older labels use a similar set under different
headings. Section 17, Patient Counseling Information, tells the prescriber what to discuss
with the patient. It is addressed to the prescriber, so the corpus treats it as a prescriber
section.

\subsection{Readability of patient medicine information}

Studies of Medication Guides have measured small sets by hand. \citet{wolf2006} reviewed all
40 guides issued at the time and found an average reading level of grade 11 to 12 by Lexile
score; none met the federal recommendation of a sixth- to eighth-grade level. A later study
assessed 185 guides, of which seven were at or below grade 8, and tested comprehension with
449 adults \citep{wolf2012}. \citet{jones2021} examined the guides of 53 new molecular entities
approved in 2011, 2013, 2015 and 2017. By Flesch-Kincaid the mean grade was 8.4 and 77\% of
the guides met a grade-8 target; by the Fog index and the Fry graph, 11\% and 9\% did. Other work compared 33 guides with Wikipedia articles
\citep{candelario2017}, assessed patient materials of risk management programs
\citep{chan2018}, compared professional and patient information for 16 antifungal medicines
across four regions \citep{moore2023}, and tested leaflets for antiseizure medicines in the
United Kingdom \citep{noble2022}. A review of 22 leaflet studies noted both the small samples
and the limits of formulas \citep{pires2015}. One recent study asked language models to rewrite
agency communications to a lower grade \citep{sridharan2024}.

Two points from this literature shape the corpus. First, readability formulas disagree.
\citet{wang2013} applied several formulas to the same health materials and found differences
of up to five grades on the same text, depending on the formula and on how the text is
prepared. They judged SMOG best suited to health information; \citet{redish2000} lists what formulas cannot see at all, such as layout and
whether the content answers the reader's question. PlainPairs therefore reports Flesch-Kincaid
and SMOG together and never one alone. Second, instruments such as the Patient Education
Materials Assessment Tool \citep{shoemaker2014} rate understandability and actionability by
hand, including layout and visual aids. The corpus's wording counts are informed by some of
its text items, but they are not scores on that instrument.

\subsection{Corpora that pair expert and lay medical text}

Table~\ref{tab:resources} lists the resources closest to PlainPairs. In most of them the plain
side is a research summary: Cochrane plain-language summaries \citep{devaraj2021}, author lay
summaries of journal abstracts \citep{guo2024}, or lay summaries of journal articles
\citep{goldsack2022}. In PLABA the plain side was written for the corpus by trained annotators
\citep{attal2023}, and Med-EASi annotates short snippets with edit operations
\citep{basu2023}. \citet{cao2020} aligned the professional and consumer versions of one
publisher's medical manual. The Spanish CLARA-MeD corpus is the closest in kind: it includes
official pairs of professional product information and patient leaflets from the Spanish
medicines agency, as well as cancer summaries \citep{campillos2022}.

For U.S.\ labels, the nearest prior work is that of \citet{meyer2023}. They took 5,357
prescription labels from the September 2019 edition of DailyMed, of which 1,437 had a
Medication Guide, and trained a summarization model to generate a guide from the prescribing
information. To prepare the training pairs they cut each guide at six of its standard
questions and aligned each answer to the label in three ways: with the whole label, with label
sections chosen by hand from domain knowledge, and with the single most similar section by
several text-similarity measures, of which BM25 worked best. The alignment in PlainPairs rests
on the same two ideas, a mapping from headings to sections and a similarity score
(Section~\ref{sec:align}). Their goal was text generation for regulatory writing, evaluated
with overlap metrics, and their data statement points to DailyMed itself and not to a released
set of pairs. PlainPairs differs in purpose and in what it ships: all labels in scope in a
current export, patient package inserts as well as guides, ten heading types, a hand check of
the mapping, reading-level and wording measures on both sides, and public files with the
rules that built them.

\begin{table}[t]
\centering\small
\caption{Resources that pair expert and lay medical text, as their papers or data cards
describe them.}
\label{tab:resources}
\begin{tabularx}{\textwidth}{@{}L{3.2cm}YL{2.9cm}L{1.6cm}@{}}
\toprule
Resource & Expert side and plain side & Size & Language \\
\midrule
PLABA \citep{attal2023} & PubMed abstracts; adaptations written for the corpus & 750 abstracts, 7,643 sentence pairs & English \\
Med-EASi \citep{basu2023} & Short expert texts; simple versions with edit annotations & 1,979 pairs & English \\
Cochrane summaries \citep{devaraj2021} & Review abstracts; the reviews' plain-language summaries & about 4,459 pairs & English \\
CELLS \citep{guo2024} & Journal abstracts; author lay summaries & 62,886 pairs & English \\
eLife and PLOS \citep{goldsack2022} & Journal articles; lay summaries & about 32,000 pairs & English \\
MSD Manual \citep{cao2020} & Professional and consumer versions of one manual & 675 aligned test pairs & English \\
CLARA-MeD \citep{campillos2022} & Professional product information and leaflets, cancer summaries, trial texts & 24,298 document pairs, 3,800 sentence pairs & Spanish \\
\citet{meyer2023} & Prescribing information; Medication Guides (2019 snapshot) & 1,437 label pairs; no public data found & English \\
\midrule
PlainPairs 0.1.0 & Prescribing information; Medication Guides and patient package inserts & 35,177 labels, 250,829 segment pairs & English \\
PlainPairs 0.1.0, \f{nci} & Cancer summaries for professionals; for patients & 756 summaries, 634 pairs & English, Spanish \\
\bottomrule
\end{tabularx}
\end{table}

% ---------------------------------------------------------------------------------------
\section{Sources and scope}
\label{sec:sources}

This section names the sources, the rules that decide which labels are in the corpus, and the
terms under which the text is released.

\subsection{Sources}

The corpus has three sources, read on the dates in Table~\ref{tab:sources}.

\textbf{openFDA drug labels.} The Food and Drug Administration publishes the labels it holds
as weekly bulk files through openFDA \citep{kasshout2016,openfda_terms}. Each record is the
current version of one label, with each section in its own field as a single string. The
export of 13 November 2026 holds 264,748 labels in 14 files. The corpus reads every export in
full, because the source states that any record may change in any update.

\textbf{DailyMed.} The National Library of Medicine publishes the same labels as Structured
Product Labeling XML \citep{dailymed}. The XML keeps what the bulk files flatten: paragraphs,
list items, headings and tables. The corpus downloads the full release of human prescription
labels (6 files, 18.1~GB, 55,148 documents for 52,840 label identifiers) and reads only the
XML documents.

\textbf{The National Cancer Institute.} The institute publishes cancer information summaries
in a version for health professionals and a version for patients, in English and Spanish. Its
reuse policy states that the text of its products is free of copyright and may be reused with
credit \citep{nci_reuse}. The corpus reads each summary page once.

Two small official lists add status flags: the National Drug Code directory, and DailyMed's
list of the labels it serves. The flags describe the lists and not the market. A label on
neither list may belong to a product that is no longer sold, or to one that the lists do not
cover.

\begin{table}[t]
\centering\small
\caption{Sources of version 0.1.0 and the dates they were read.}
\label{tab:sources}
\begin{tabular}{@{}llr@{}}
\toprule
Source & Used for & Date \\
\midrule
openFDA drug label export & every label and its text & 13 Nov 2026 \\
openFDA NDC directory export & the flag \f{listed_in_ndc_directory} & 13 Nov 2026 \\
DailyMed full release, human prescription labels & text structure of \f{current} labels & 13 Nov 2026 \\
DailyMed list of served labels & the flag \f{served_by_dailymed} & 15 Nov 2026 \\
National Cancer Institute summary pages & the \f{nci} configuration & 15 Nov 2026 \\
\bottomrule
\end{tabular}
\end{table}

\subsection{Scope}

A label is in scope when it is in a prescription format and has a Medication Guide or a patient
package insert that is not empty after cleaning. The format is read from the sections a label
has, not from a type field, because most records lack one. Table~\ref{tab:scope} follows the
export down to the corpus. Of 264,748 labels, 35,177 are in scope: 25,796 in the newer
prescription format and 9,381 in the older one. A Medication Guide is present on 22,872 of
them and a patient package insert on 12,519; 214 have both.

One hundred labels that the source types as over-the-counter products also carry a guide or an
insert. They are out of scope, and the release lists them in a table of their own so that the
choice can be checked.

\begin{table}[t]
\centering\small
\caption{From the export of 13 November 2026 to the corpus.}
\label{tab:scope}
\begin{tabular}{@{}lr@{}}
\toprule
 & Labels \\
\midrule
Labels in the export & 264,748 \\
In a prescription format, with a Medication Guide or a patient package insert & 35,177 \\
\quad newer prescription format / older format & 25,796 / 9,381 \\
\quad with a Medication Guide / with a patient package insert / with both & 22,872 / 12,519 / 214 \\
\quad with Instructions for Use & 2,059 \\
\quad with the source's product block (\f{openfda}) & 15,022 \\
\quad showing a copyright mark in any text field & 3,781 \\
Configuration \f{current}: served by DailyMed in the same version & 19,778 \\
Configuration \f{unlisted}: every other label & 15,399 \\
\midrule
Distinct Medication Guides (after removing repeats) & 12,006 \\
Distinct patient package inserts & 7,769 \\
Distinct patient texts of all four kinds & 21,416 \\
\bottomrule
\end{tabular}
\end{table}

\subsection{Rights and the text switch}
\label{sec:rights}

openFDA states that its content is in the public domain under a CC0 dedication unless
otherwise noted \citep{openfda_terms}. The labels themselves are written by the companies that
market each medicine, and 3,781 of the 35,177 labels in scope (10.7\%) show a copyright sign,
the word ``copyright'' or ``all rights reserved'' somewhere in their text. I do not know
whether those marks limit reuse of label text, and the corpus does not try to settle the
question. It is built in two layers instead. The record layer holds identifiers, dates,
hashes, lengths, measures, counts and alignment scores, and no text. The text layer holds the
cleaned texts. A switch decides which text rows a release carries: all, only those of labels
with no mark, or none. Version 0.1.0 uses the middle setting, which holds back 68,156 text
rows of marked labels while keeping their record rows. A command in the software rebuilds any
label's text from the source's own files and confirms it against the stored hash, so the text
layer can be reproduced by anyone who downloads the same export.

The release also has a removal route. A company or any reader can ask that a text be taken
out; the text row is then removed in the next version and the record row stays.

% ---------------------------------------------------------------------------------------
\section{Building the corpus}
\label{sec:method}

\begin{figure}[t]
\centering
\begin{tikzpicture}[
  node distance=5mm and 5mm,
  box/.style={draw, rounded corners=2pt, align=center, font=\footnotesize, inner sep=4pt, minimum height=11mm, text width=27mm},
  src/.style={box, fill=black!6},
  arr/.style={-{Stealth[length=2mm]}, semithick}]
\node[src] (fda) {openFDA export\\264,748 labels};
\node[box, right=of fda] (scope) {Parse and scope\\35,177 labels};
\node[box, right=of scope] (dup) {Identity and repeats\\21,416 patient texts};
\node[box, right=of dup] (struct) {Join to label XML\\\texttt{current} 19,778\\\texttt{unlisted} 15,399};
\node[src, below=of struct] (dm) {DailyMed XML\\55,148 documents};
\node[box, below=of dup] (meas) {Measures and\\wording counts};
\node[box, below=of scope] (align) {Cut at headings,\\map and score\\250,829 segments};
\node[box, below=of fda] (pack) {Release pack\\record layer, text layer,\\checks};
\node[src, below=of meas] (nci) {NCI summary pages\\756 summaries};
\draw[arr] (fda) -- (scope);
\draw[arr] (scope) -- (dup);
\draw[arr] (dup) -- (struct);
\draw[arr] (dm) -- (struct);
\draw[arr] (struct.south west) -- (meas.north east);
\draw[arr] (meas) -- (align);
\draw[arr] (align) -- (pack);
\draw[arr] (nci) -- (meas);
\end{tikzpicture}
\caption{The build pipeline of version 0.1.0.}
\label{fig:pipeline}
\end{figure}
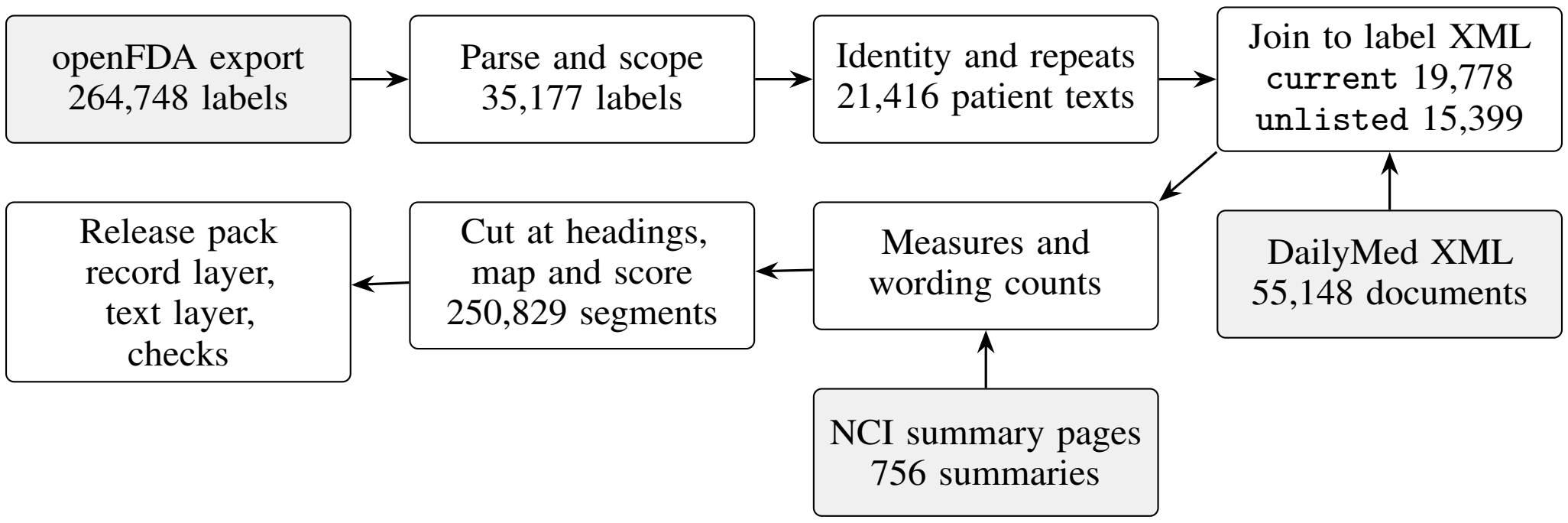

This section follows the pipeline in Figure~\ref{fig:pipeline} from the raw export to the
release. All steps are fixed rules. No language model wrote, labeled or scored any text in the
corpus.

\subsection{Parsing, identity and repeats}

Each label is read from the bulk files and becomes one row. Every section is cleaned of markup
remains and stray spacing with all words kept. Tables arrive as HTML and are stored apart from
the running text, and the copies of tables that the source repeats inside a section's text are
cut before any measure is taken.

A label is identified by its set identifier, which is stable across versions. Naming the
medicine is harder, because only 15,022 of the 35,177 labels carry the source's product block
with generic names. For the rest, the corpus borrows a medicine key from a label that carries
an identical patient text and does have generic names, and otherwise leaves it empty. As a
result, 16,122 labels have no medicine key, and the count of distinct medicines (1,713) is a
lower bound.

Many labels share a patient text, because generic products copy the guide of the original
product. Each patient text gets three hashes: exact, loose (ignoring case, punctuation and
symbols), and core (loose, without the closing lines that name the maker and the revision
date). Texts are grouped by the core hash. Reading levels are always reported over distinct
texts, so that a guide used by 40 labels counts once.

\subsection{Structure and the two configurations}
\label{sec:structure}

The bulk files store each section as one string. A bulleted list then reads as one long
sentence, and a heading runs into the paragraph below it. This matters for every measure that
depends on sentence length. To recover the layout, each label is looked up in the DailyMed
release by set identifier and version. When the release holds the same version, the words of
each section in the XML are compared with the words of the bulk file. A section matches when
the two token sequences are identical, or when the Dice overlap \citep{dice1945} of their
token bags,
\begin{equation}
  D(A,B) = \frac{2\,|A \cap B|}{|A| + |B|},
  \label{eq:dice}
\end{equation}
is 0.98 or more. A third class covers sections where the bulk file repeats a subsection that
the XML holds once.

For matched labels the XML is read into blocks: titles, headings, paragraphs, list items,
cells of tables that hold prose, and captions. Every paragraph, item, prose cell and caption
is cut into sentences, and every block ends a sentence. Titles, headings and cells of data
tables are not sentences and are not measured.

This gives two label configurations that are never pooled:
\begin{itemize}
  \item \f{current} (19,778 labels): the DailyMed release holds the label in the same version,
  and every patient text has a matching section in the XML. These labels carry measures from
  structure next to their measures from the flattened text.
  \item \f{unlisted} (15,399 labels): every other label, measured from the flattened text only.
  Of these, 15,122 are not in the DailyMed release, 268 are there in another version, and 9
  have a patient text that the XML lacks or words differently.
\end{itemize}
The two configurations hold different labels measured in different ways. A difference between
them is not a measured change, and no figure in this paper combines them except counts of rows.

\subsection{Reading-level measures}
\label{sec:measures}

For each text, with $W$ words, $S$ sentences, $Y$ syllables and $P$ words of three or more
syllables, the corpus stores the Flesch-Kincaid grade \citep{kincaid1975},
\begin{equation}
  \mathrm{FK} = 0.39\,\frac{W}{S} + 11.8\,\frac{Y}{W} - 15.59,
  \label{eq:fk}
\end{equation}
the SMOG grade \citep{mclaughlin1969} in the form most software uses,
\begin{equation}
  \mathrm{SMOG} = 1.0430\,\sqrt{P \cdot \frac{30}{S}} + 3.1291,
  \label{eq:smog}
\end{equation}
and the Flesch reading ease score \citep{flesch1948}. Syllables are counted by rule. SMOG was
designed for texts of at least 30 sentences, and shorter texts carry a flag. The corpus also
stores the share of sentences longer than 20 words and counts of numbers and percentages
written in digits.

For Spanish text the corpus stores the Szigriszt-Pazos perspicuity index,
\begin{equation}
  \mathrm{SP} = 206.835 - 62.3\,\frac{Y}{W} - \frac{W}{S},
  \label{eq:sp}
\end{equation}
read on the INFLESZ scale, which was validated on texts written for patients
\citep{barrio2008}. This index is an ease score on which higher is easier. It is not a school
grade and is never set beside an English grade. I treat it as provisional, because the Spanish
syllable rule was tested on only 46 words.

Grades are comparable between texts measured the same way. They are not the grade a reader
meets on the printed page, which has a layout that neither the bulk files nor the XML fully
keep.

\subsection{Alignment of patient segments to prescriber sections}
\label{sec:align}

The alignment has three steps: cut, map and score. The map and score steps correspond to what
\citet{meyer2023} called manual and heuristic alignment; here both are kept for every segment,
so that a user can compare them.

\textbf{Cut.} Each guide or insert is cut at its standard headings, found by their wording.
There are ten heading keys, listed in Table~\ref{tab:sheet}, which cover the questions set by
regulation \citep{cfr208} and the common additions. A heading that is only cited inside
running text is skipped.

\textbf{Map.} Each heading key is mapped to the prescriber sections that cover the same ground.
The segment on what the medicine is maps to Indications and Usage, the one on who should not
take it to Contraindications, how to take it to Dosage and Administration, storage to How
Supplied, and ingredients to Description. The most important information maps to the Boxed
Warning and to Warnings and Precautions, and side effects to Adverse Reactions and Warnings
and Precautions. Two keys map to three sections each: what to tell the prescriber before
taking the medicine, and what to avoid while taking it. The segment of general information has
no counterpart named by the regulation, and I mapped it to Patient Counseling Information by
my own choice.

\textbf{Score.} Each segment is also compared with every prescriber section of its label,
whatever the mapping says. A word $t$ that occurs $c$ times in a text gets the weight
\begin{equation}
  w_t = (1 + \ln c)\left(\ln\frac{1+N}{1+n_t} + 1\right),
  \label{eq:tfidf}
\end{equation}
where $N$ is the number of texts in the corpus and $n_t$ the number that contain $t$
\citep{salton1988}. Words found in more than half of one label's own sections, which are
mostly its product names, are left out for that label. The similarity of a segment $a$ and a
section $b$ is the cosine of their weight vectors,
\begin{equation}
  \mathrm{sim}(a,b) = \frac{\sum_t w_{t,a}\,w_{t,b}}{\sqrt{\sum_t w_{t,a}^2}\,\sqrt{\sum_t w_{t,b}^2}}.
  \label{eq:cos}
\end{equation}
Each segment pair row keeps the mapped sections that the label has, the most similar section,
both similarities, and whether the most similar section is one of the mapped ones.

\subsection{Rule-based counts of wording}
\label{sec:codes}

The corpus counts wording features that the plain-language literature treats as signs of text
written for a lay reader, or of text that is hard for one. The first release carries nine:
words and sentences that address the reader as ``you''; instructions given as commands, and the
number of distinct actions they name; percentages; risks given as ``1 in $N$''; frequencies;
doses; sentences that say whom to contact; and, for \f{current} labels, sentences over 20 and
over 40 words read from structure. Each is a fixed rule over one cleaned text, stored as a
count.

Seven further codes were written and are held back as experimental, because a hand check
found too many errors (Section~\ref{sec:validation}). They include statements of risk with a
number, vague frequency words, and terms explained in brackets.

\subsection{The cancer summaries}

The National Cancer Institute's summaries form a third configuration, \f{nci}, kept in its own
tables. The list of summaries comes from the institute's sitemap. Each page is read once, at
least five seconds after the last request. The body is read into the same kinds of blocks as
the labels, and the same measures are applied; the wording counts are computed for English
summaries only.

Several parts of each page are left out and counted: 1,562 images and figures, which the
institute uses under permissions that cover its own pages; 763 tables that credit another
publisher; the reference lists (68,229 items) and the citation marks in the text; navigation;
and the closing sections about the summary itself. Because parts are left out, a row of this
configuration is not one of the institute's summaries and should not be presented as one.
Every row carries the page's address, the date it was read, the institute's update date and a
credit line.

Pairs come only from the institute's own links. A version pair joins a health-professional
summary with the patient summary that its page links to. A language pair joins a summary with
its page in the other language. The configuration holds 267 version pairs (131 English, 136
Spanish) and 367 language pairs.

\subsection{Release, checks and updates}

A release is written as one folder per version with a manifest that gives every file's row
count, size and SHA-256 hash, the date of each source and the setting of the text switch. A
check that runs on the folder alone fails when a file is missing or changed, when a text row
of a marked label is present, when an experimental code appears, when measures from structure
sit on an \f{unlisted} row, or when a row of the cancer configuration lacks its credit line.

The software includes an update routine that reads a new export, rebuilds the tables and
writes a table of changes: labels that are new, withdrawn, re-versioned, or changed in text.
At the time of writing this routine has been tested on synthetic exports only, because no
export newer than the one in version 0.1.0 had been published. The release therefore describes
one export, and I make no claim yet about how the corpus behaves over time.

% ---------------------------------------------------------------------------------------
\section{Data records}
\label{sec:records}

This section describes the files of version 0.1.0 and their main fields.

The release has 16 Parquet files with 1,409,294 rows in 563~MB (Table~\ref{tab:files}). It is
published on Hugging Face \citep{plainpairs_data}, where each table is a configuration of its
own, and mirrored on Kaggle. There is no default configuration, so a user names the one table
wanted and nothing loads two label configurations as one table. The software and documents are
archived on Zenodo \citep{plainpairs_code}. The release is meant to be findable and reusable
in the sense of the FAIR principles \citep{wilkinson2016}: each version has a tag, a manifest
with hashes, stated terms for every source, and a data card written along the lines of a
datasheet \citep{gebru2021}.

\begin{table}[t]
\centering\small
\caption{Files of version 0.1.0.}
\label{tab:files}
\begin{tabular}{@{}lllrr@{}}
\toprule
Configuration & Table & Layer & Rows & MB \\
\midrule
\f{current} & \f{labels} & record & 19,778 & 20.2 \\
 & \f{patient_texts} & record & 12,857 & 2.5 \\
 & \f{section_pairs} & record & 143,885 & 5.1 \\
 & \f{codes} & record & 221,072 & 2.9 \\
 & \f{text_prescriber} & text & 260,290 & 242.6 \\
 & \f{text_patient} & text & 33,003 & 41.4 \\
\f{unlisted} & \f{labels} & record & 15,399 & 11.0 \\
 & \f{patient_texts} & record & 8,559 & 1.3 \\
 & \f{section_pairs} & record & 106,944 & 3.9 \\
 & \f{codes} & record & 171,088 & 1.8 \\
 & \f{text_prescriber} & text & 196,019 & 184.6 \\
 & \f{text_patient} & text & 23,516 & 29.2 \\
\f{nci} & \f{summaries} & record & 756 & 0.1 \\
 & \f{pairs} & record & 634 & 0.0 \\
 & \f{text} & text & 195,394 & 16.3 \\
(a list) & \f{excluded_otc} & record & 100 & 0.0 \\
\midrule
All & 16 files & & 1,409,294 & 563.0 \\
\bottomrule
\end{tabular}
\end{table}

\textbf{\f{labels}} has one row per label: its identifiers and dates, format, the source's
product fields where they exist, the medicine key and where it came from, the two status
flags, the configuration, flags for each kind of text and for a copyright mark, and, for each
of 11 prescriber sections and 4 patient texts, a hash and a set of measures. \f{current}
labels have a second set of measures from structure.

\textbf{\f{patient_texts}} has one row per distinct patient text: its kind, the number of
labels that share it, the number of exact and loose variants, whether those labels name more
than one maker, the standard headings found in it, and its measures.

\textbf{\f{section_pairs}} has one row per segment of a guide or insert on a label: the heading
key, the position of the segment in the text, its measures, the mapped sections, the most
similar section, the similarities, and \f{rule_status}, which says whether the heading's
mapping passed the hand check.

\textbf{\f{codes}} has one row per text or section with the wording counts of
Section~\ref{sec:codes}.

\textbf{\f{text_prescriber}} and \textbf{\f{text_patient}} hold the cleaned texts, one row per
label and section, with tables as separate rows and a hash for each row.

The \f{nci} tables follow the same pattern: \f{summaries} with identity, credit and measures,
\f{pairs} with the two addresses of each pair, and \f{text} with one row per kept block.
Appendix~\ref{app:fields} lists the main fields.

Table~\ref{tab:example} shows what one row of \f{section_pairs} joins. The two passages in it
are invented for this paper in the style of the two kinds of text. They are not taken from any
label and say nothing about any medicine.

\begin{table}[t]
\centering\small
\caption{An invented example of a segment pair; the wording and the values are illustrative only.}
\label{tab:example}
\begin{tabularx}{\textwidth}{@{}L{3.6cm}Y@{}}
\toprule
Patient segment, under the heading on how to take the medicine & ``Take [medicine] exactly as your healthcare provider tells you. Take it one time each day, with or without food. Do not change your dose or stop taking [medicine] without talking to your healthcare provider.'' \\
\addlinespace
Prescriber section, Dosage and Administration & ``The recommended dosage is 10 mg orally once daily, with or without food. The dosage may be increased to 20 mg once daily after 4 weeks, based on clinical response and tolerability.'' \\
\addlinespace
Fields of the row & \f{mapped_sections}: \f{dosage_and_administration} \newline \f{best_section}: \f{dosage_and_administration} \newline \f{best_is_mapped}: true \newline \f{best_similarity}: 0.21 \newline \f{rule_status}: \f{rule_released} \\
\addlinespace
Wording counts, patient side & 3 sentences; 3 address the reader; 3 give a command; 0 percentages; 0 doses \\
Wording counts, prescriber side & 2 sentences; 0 address the reader; 0 give a command; 0 percentages; 2 doses \\
\bottomrule
\end{tabularx}
\end{table}

% ---------------------------------------------------------------------------------------
\section{Technical validation}
\label{sec:validation}

This section reports four checks: that the XML and the bulk files hold the same words, how
much the measurement basis changes reading levels, how the wording counts were checked, and
how the alignment was checked. The last two were done by one reader, and I say what that
limits.

\subsection{The same words in both sources}

Measures from structure are only meaningful if the XML holds the same text as the bulk file.
Of the 12,857 distinct patient texts in \f{current}, 12,848 match word for word, 6 match at a
Dice overlap of 0.98 or more, and 3 belong to the class where the bulk file repeats a
subsection. Labels where a patient text did not match are in \f{unlisted} by construction.
Among prescriber sections, the bulk file repeats a subsection in 1.6\% to 3.3\% of cases,
depending on the section. Word counts taken from the flattened text are too high for those
sections, and the measures from structure are not affected.

\subsection{Structure against flattened text}
\label{sec:basis}

\begin{figure}[t]
\centering
\includegraphics[width=\textwidth]{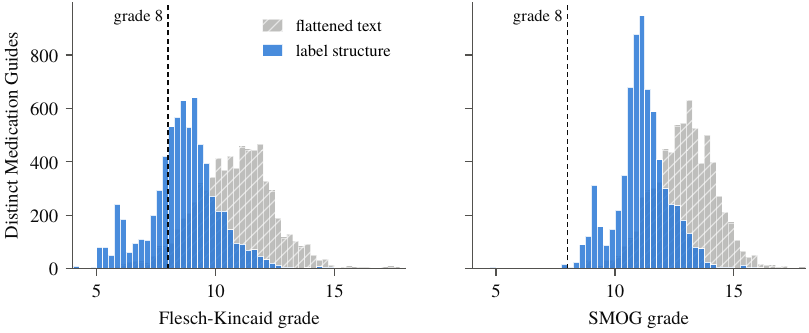}
\caption{Reading level of 7,088 distinct Medication Guides in \f{current}, on two bases.}
\label{fig:basis}
\end{figure}

Figure~\ref{fig:basis} compares the two bases on the same 7,088 distinct Medication Guides.
Measured on structure, the median guide has 134 sentences; measured on the flattened text it
has 98, because the 56 list items of the median guide are joined into fewer, longer units.
The Flesch-Kincaid grade from the flattened text is higher in 98.9\% of guides, by a median of
1.94 grades (middle half 1.17 to 2.76), and the SMOG grade is higher by a median of 1.63. For
patient package inserts the median differences are 1.44 and 1.19 grades.

The size of this gap is a finding about the source as much as a check of the corpus. A study
that computes readability from the bulk files alone will place the median guide near grade 11
by Flesch-Kincaid, and one that reads the label's structure will place it near grade 9.
Neither figure is wrong, but they answer different questions, and they should not be compared
with each other.

\subsection{Check of the wording counts}

Each code was checked against its own definition on items drawn with fixed seeds, half from
patient texts and half from prescriber sections: 10 hits and 10 misses for most codes, and 20
of each for the five codes I doubted from the start. A hit is right when the unit is what the
code says it is. A miss is right when the code was right to leave the unit out. A code is
released when at least 90\% of its hits and 90\% of its misses are right.

Table~\ref{tab:codes} gives the results for the released codes. Two rules were corrected after
the first reading: the ``1 in $N$'' rule had counted packaging counts, and the rule for whom to
contact had counted sentences that tell the prescriber to inform the patient. Seven codes
failed and are not released. For example, long sentences read from the flattened text were
right in only 8 of 20 hits, because list items had run together, while the same code read
from structure was right in 19 of 20.

This was a development check and not a validation. One reader made every judgment, there was
no second coder, and no agreement statistic can be reported. It shows where a rule fails often
enough to be withheld. It does not show that a released count measures what a human rater of
plain language would call the same feature.

\begin{table}[t]
\centering\small
\caption{Hand check of the released wording counts.}
\label{tab:codes}
\begin{tabular}{@{}lll@{}}
\toprule
Code & Hits right & Misses right \\
\midrule
Direct address (``you'') & 10 of 10 & 10 of 10 \\
Instructions as commands & 10 of 10 & 10 of 10 \\
Percentages & 10 of 10 & 10 of 10 \\
Risk as ``1 in $N$'' & 9 of 10; 9 of 9 after the fix & 10 of 10 \\
Frequencies & 9 of 10 & 10 of 10 \\
Doses & 10 of 10 & 10 of 10 \\
Whom to contact & 36 of 40; 36 of 36 after the fix & not reported \\
Long sentences, from structure & 19 of 20 & 20 of 20 \\
\bottomrule
\end{tabular}
\end{table}

\subsection{Check of the alignment}
\label{sec:sheet}

To check the mapping of headings to sections, I drew 100 segment pairs with a fixed seed, 10
per heading key, never two from the same distinct patient text, and judged each one blind to
its similarity score. The sheet asked one question: does this prescriber section cover the
same topic as this patient segment (yes, partly, or no)? It did not ask whether the content is
medically right. I set the rule in advance that a heading keeps its mapping when at least 80\%
of its rows are ``yes'' or ``partly''.

\begin{figure}[t]
\centering
\includegraphics[width=\textwidth]{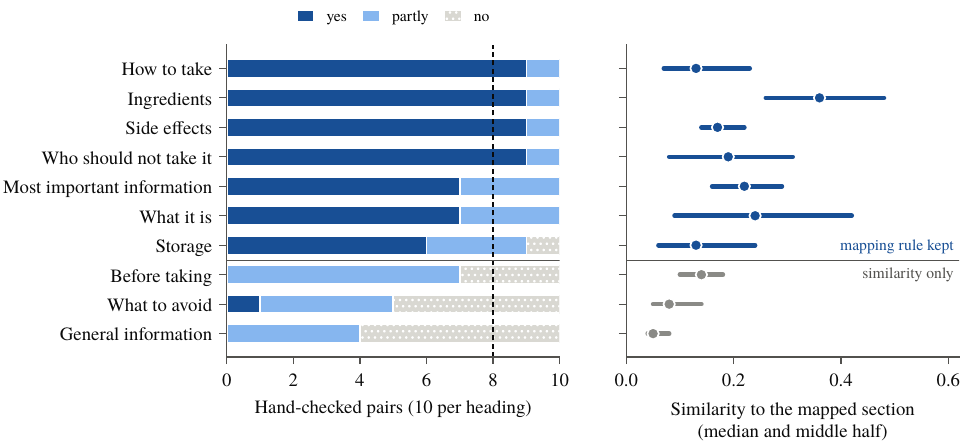}
\caption{Alignment by heading key: the hand check, and similarity to the mapped section in \f{current}.}
\label{fig:alignment}
\end{figure}

Seven headings passed and three did not (Table~\ref{tab:sheet}, Figure~\ref{fig:alignment}).
The three that failed are the two whose mapping points to three sections at once and the one
whose mapping was my own choice. Their pairs are released with similarity scores only and
are flagged in \f{rule_status}. Across the two label configurations, 184,125 segment rows carry
a mapping that passed and 66,704 carry the flag.

Ten rows per heading give wide intervals. A heading at 10 of 10 has a 95\% Wilson interval
\citep{wilson1927} from 72\% to 100\%, so the check shows that a mapping is not badly wrong,
and it does not show that it is right in nearly all cases. The check was also made by one
coder. I plan a second pass over the same 100 pairs after a gap of two weeks, to measure my
own consistency, and the result will be reported with the next version.

The similarity scores give a second, independent view. They are low in absolute terms, with
median values from 0.05 to 0.36 by heading, which is expected when a short lay passage is
compared with a long technical section in different vocabulary. The mapped section is the most
similar section of the label for 54\% to 86\% of segments, depending on the heading. Two of the
three headings that failed the hand check have the lowest median similarity to their mapped
sections (0.05 and 0.08). The third, at 0.14, sits among headings that passed, so the
similarity score alone would not have separated them.

\begin{table}[t]
\centering\footnotesize
\caption{Hand check of the alignment: 100 segment pairs, 10 per heading key, one coder.}
\label{tab:sheet}
\begin{tabularx}{\textwidth}{@{}L{3.3cm}Yrrrll@{}}
\toprule
Heading key & Mapped prescriber sections & Yes & Partly & No & Yes or partly (95\% interval) & Result \\
\midrule
How to take & Dosage and Administration & 9 & 1 & 0 & 100\% (72 to 100) & kept \\
Ingredients & Description & 9 & 1 & 0 & 100\% (72 to 100) & kept \\
Side effects & Adverse Reactions; Warnings and Precautions & 9 & 1 & 0 & 100\% (72 to 100) & kept \\
Who should not take it & Contraindications & 9 & 1 & 0 & 100\% (72 to 100) & kept \\
Most important information & Boxed Warning; Warnings and Precautions & 7 & 3 & 0 & 100\% (72 to 100) & kept \\
What it is & Indications and Usage & 7 & 3 & 0 & 100\% (72 to 100) & kept \\
Storage & How Supplied and Storage & 6 & 3 & 1 & 90\% (60 to 98) & kept \\
Before taking & Warnings and Precautions; Use in Specific Populations; Drug Interactions & 0 & 7 & 3 & 70\% (40 to 89) & similarity only \\
What to avoid & Warnings and Precautions; Drug Interactions; Patient Counseling Information & 1 & 4 & 5 & 50\% (24 to 76) & similarity only \\
General information & Patient Counseling Information (the project's choice) & 0 & 4 & 6 & 40\% (17 to 69) & similarity only \\
\bottomrule
\end{tabularx}
\end{table}

\subsection{Reproducibility checks}

Three further checks are part of every build. The text layer is rebuilt from the record layer
and the source's files, and every row is confirmed by its hash. The cancer configuration was
rebuilt from the saved pages, and all 756 summaries came out the same. After publication, the
public files were compared with the release folder by hash without credentials, and two small
tables were loaded from a clean environment with the row counts of the manifest.

% ---------------------------------------------------------------------------------------
\section{What the first release shows}
\label{sec:findings}

This section reports five descriptive results. All figures are for one export and are given
per configuration.

\subsection{Reading levels of patient texts}

Table~\ref{tab:levels} gives reading levels over distinct texts. In \f{current}, measured on
structure, the median Medication Guide reads at grade 8.67 by Flesch-Kincaid (middle half 7.88
to 9.48) and at grade 11.10 by SMOG. By Flesch-Kincaid, 71.9\% of guides are above grade 8; by
SMOG, 99.2\% are. Patient package inserts are close to the guides: grade 8.58 and 11.21.

Whether Medication Guides meet a grade-8 target therefore depends on the formula. On
Flesch-Kincaid, more than a quarter do and the median is within a grade of the target. On
SMOG, which is built on words of three or more syllables and is calibrated to full
comprehension of a text \citep{wang2013}, almost none do. \citet{jones2021} found the same
split on 53 guides, with a mean Flesch-Kincaid grade of 8.4 that is close to the median of
8.67 measured here on 7,182. The two formulas were
calibrated against different criteria, and medicine names and medical terms are long words
that a guide cannot avoid, so I do not read the SMOG result as evidence that the guides are
poorly written. It does show that a single formula should not be used to say that these texts
are or are not readable.

Instructions for Use read lower than either: grade 6.68 by Flesch-Kincaid and 9.75 by SMOG,
which fits texts made of short numbered steps.

\begin{table}[t]
\centering\small
\caption{Reading levels of distinct patient texts, per configuration and measurement basis.}
\label{tab:levels}
\begin{tabular}{@{}llrrrrr@{}}
\toprule
 & & & \multicolumn{2}{c}{Flesch-Kincaid} & \multicolumn{2}{c}{SMOG} \\
\cmidrule(lr){4-5}\cmidrule(lr){6-7}
Text & Configuration and basis & Texts & Median & Above 8 & Median & Above 8 \\
\midrule
Medication Guide & \f{current}, structure & 7,182 & 8.67 & 71.9\% & 11.10 & 99.2\% \\
 & \f{current}, flattened text & 7,182 & 10.86 & 95.0\% & 12.94 & 99.5\% \\
 & \f{unlisted}, flattened text & 4,824 & 10.70 & 93.1\% & 12.85 & 99.5\% \\
Patient package insert & \f{current}, structure & 4,576 & 8.58 & 66.9\% & 11.21 & 99.5\% \\
 & \f{current}, flattened text & 4,576 & 10.03 & 91.4\% & 12.46 & 99.6\% \\
 & \f{unlisted}, flattened text & 3,193 & 9.72 & 87.6\% & 12.20 & 99.7\% \\
Instructions for Use & \f{current}, structure & 1,033 & 6.68 & 23.6\% & 9.75 & 94.5\% \\
\bottomrule
\end{tabular}
\end{table}

\subsection{Patient texts against prescriber sections}

\begin{figure}[t]
\centering
\includegraphics[width=\textwidth]{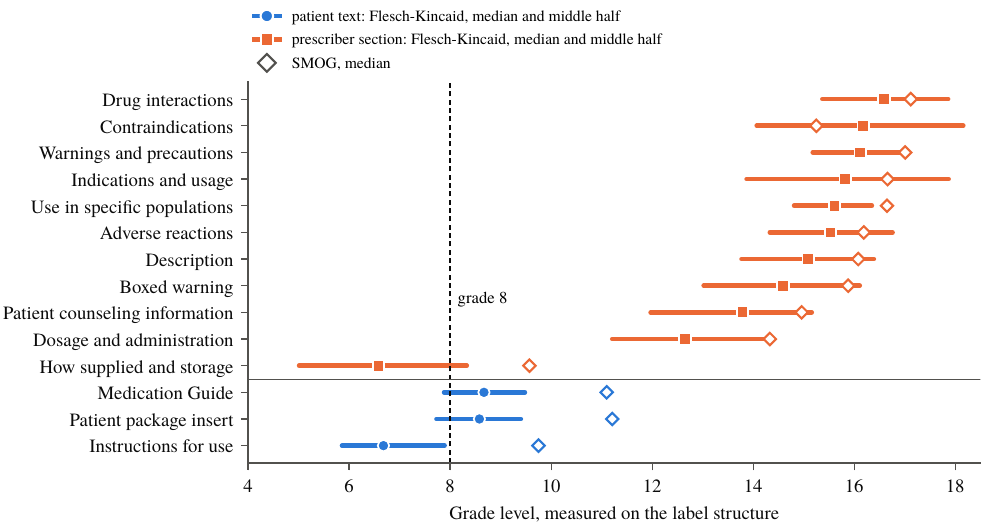}
\caption{Reading level of patient texts and prescriber sections in \f{current}, over distinct texts.}
\label{fig:sections}
\end{figure}

Figure~\ref{fig:sections} sets the patient texts beside the 11 prescriber sections, all
measured on structure in \f{current}. Ten of the 11 prescriber sections have a median
Flesch-Kincaid grade between 12.6 and 16.6. Dosage and Administration is the lowest of these
at 12.65, and Patient Counseling Information, which lists what to tell the patient, follows at
13.79. The one exception is How Supplied and Storage (6.58), which is mostly short statements
of package sizes and storage temperatures.

\begin{figure}[t]
\centering
\includegraphics[width=0.62\textwidth]{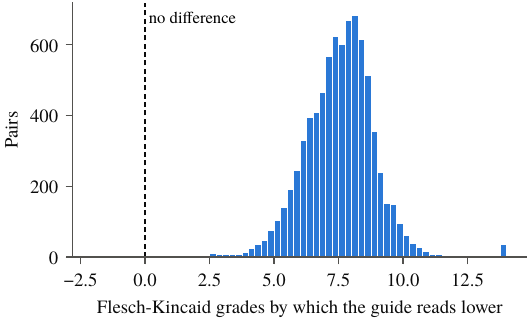}
\caption{Gap between the Warnings and Precautions section and the Medication Guide of the same label, 8,045 pairs in \f{current}.}
\label{fig:gap}
\end{figure}

The comparison within labels is sharper than the comparison of medians. I paired each
distinct Medication Guide with the Warnings and Precautions section of a label that carries it.
In 8,045 such pairs the guide has the lower Flesch-Kincaid grade in 99.9\%, with a median gap
of 7.62 grades (6.07 by SMOG). Against Dosage and Administration the guide is lower in 95.7\%
of 7,794 pairs, with a median gap of 4.44 grades. The patient text is, as intended, a much
simpler text than the section it draws on (Figure~\ref{fig:gap}). How much simpler it should
be is a question the corpus can inform and cannot answer.

\subsection{Wording}

\begin{figure}[t]
\centering
\includegraphics[width=\textwidth]{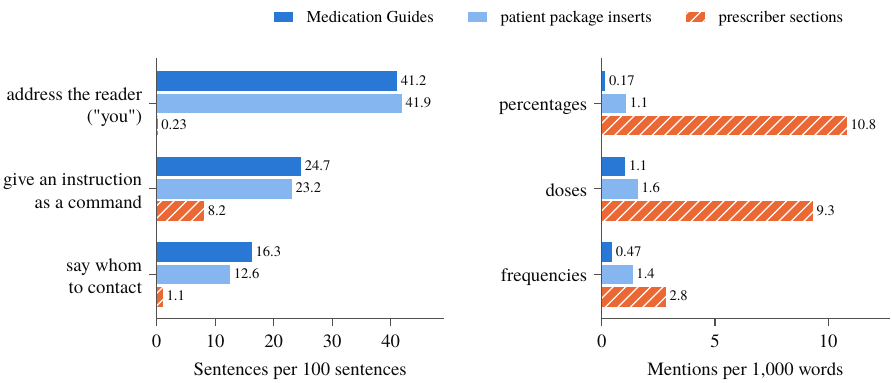}
\caption{Wording counts in distinct patient texts and prescriber sections of \f{current}.}
\label{fig:codes}
\end{figure}

The wording counts separate the two kinds of text more clearly than the grades do
(Figure~\ref{fig:codes}). In Medication Guides, 41.2 of every 100 sentences address the reader
as ``you'', 24.7 give an instruction as a command, and 16.3 say whom to contact. In prescriber
sections the same rates are 0.2, 8.2 and 1.1. Numbers run the other way. Prescriber sections
contain 10.8 percentages per 1,000 words, and the Adverse Reactions section alone 39.3;
Medication Guides contain 0.17. Only 3.3\% of guides contain any percentage at all, against
91.5\% of Adverse Reactions sections.

The near absence of numbers in the patient texts deserves attention from researchers who
study how risk is communicated. The prescriber is told how often an adverse reaction occurred
in trials. The patient is usually told that it can occur. Patient package inserts keep
somewhat more: 1.09 percentages per 1,000 words, and 15.3\% of inserts contain at least one.
The corpus cannot say which choice serves patients better, but it makes the choice visible for
every label.

\subsection{The cancer summaries}

\begin{figure}[t]
\centering
\includegraphics[width=0.92\textwidth]{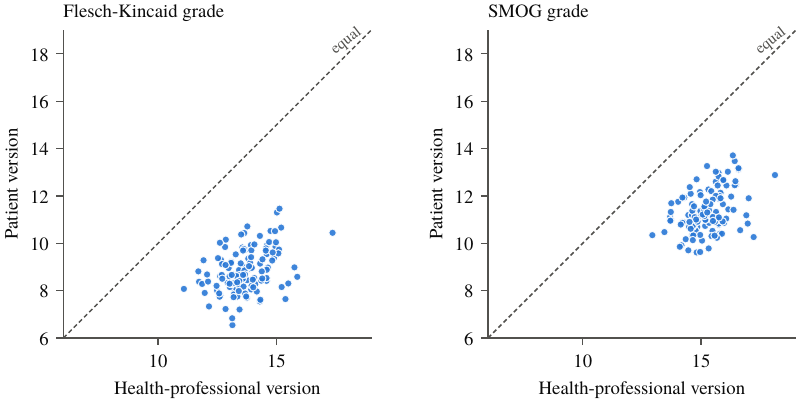}
\caption{Reading level of the 131 English version pairs of the \f{nci} configuration.}
\label{fig:nci}
\end{figure}

The cancer configuration gives the same picture from an unrelated source
(Table~\ref{tab:nci}, Figure~\ref{fig:nci}). In each of the 131 English version pairs the
patient summary reads lower than the health-professional summary on both formulas, by a median
of 4.84 grades on Flesch-Kincaid and 3.90 on SMOG, and it is about half as long (a median of
0.47 of the words). Over all English summaries, the median patient summary reads at grade 8.65
by Flesch-Kincaid and 11.11 by SMOG.

Those two medians are almost the same as the medians for Medication Guides measured on
structure (8.67 and 11.10). The two sets of texts come from different institutions, cover
different subjects and were measured by the same rules. I do not want to make too much of
one coincidence, but it suggests a level at which careful institutional writing for patients
tends to settle, and it is a level that passes one common target and misses another.

In Spanish, the patient summary has the higher (easier) Szigriszt-Pazos score in 135 of 136
version pairs, with medians of 57.7 for patient summaries and 46.9 for health-professional
summaries in the pairs. These Spanish figures are provisional (Section~\ref{sec:measures}).

\begin{table}[t]
\centering\small
\caption{The \f{nci} configuration: summaries and reading levels, each summary once.}
\label{tab:nci}
\begin{tabular}{@{}lrrrrr@{}}
\toprule
 & & & \multicolumn{2}{c}{English} & Spanish \\
\cmidrule(lr){4-5}\cmidrule(lr){6-6}
Version & English & Spanish & Flesch-Kincaid & SMOG & Szigriszt-Pazos \\
\midrule
Health professional & 242 & 220 & 13.49 & 15.08 & 47.29 \\
Patient & 147 & 147 & 8.65 & 11.11 & 57.82 \\
\midrule
Version pairs & 131 & 136 & & & \\
Language pairs & \multicolumn{2}{c}{220 professional, 147 patient} & & & \\
\bottomrule
\end{tabular}
\end{table}

\subsection{Headings, shared texts and label dates}

\begin{figure}[t]
\centering
\includegraphics[width=0.62\textwidth]{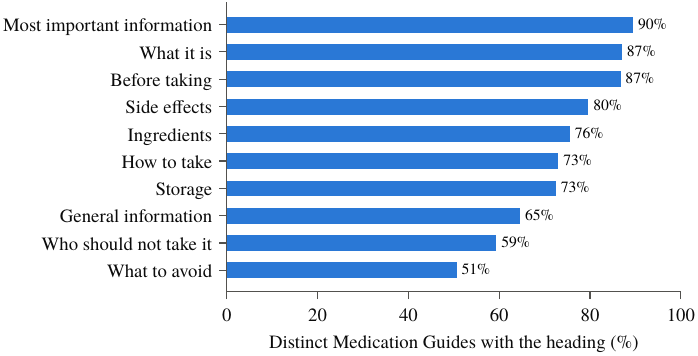}
\caption{Standard headings found in the 7,182 distinct Medication Guides of \f{current}.}
\label{fig:headings}
\end{figure}

Medication Guides do not all carry the same headings (Figure~\ref{fig:headings}). The segment
on the most important information is present in 90\% of distinct guides in \f{current}, and
the segments on what the medicine is and on what to tell the prescriber in 87\% each. The
segment on who should not take the medicine is present in 59\%, and the one on what to avoid
in 51\%. The median guide has 8 of the 10 standard headings. This is what the regulation
provides for, since a guide carries the headings relevant to its product \citep{cfr208}. A
missing heading does not mean missing content, because a guide may cover the point under
another heading, but it limits how many segment pairs each heading contributes. The shares
agree with an independent count. \citet{meyer2023} reported how often six of these questions
occur in their 1,437 guides from 2019: 87\% for the most important information, 89\% for what
the medicine is, 63\% for who should not take it, 70\% for how to take it, 52\% for what to
avoid and 77\% for side effects. The figures here for the same six (90\%, 87\%, 59\%, 73\%,
51\% and 80\%) are within four points of each.

Two further properties of the label data affect how the corpus should be analyzed. First, texts are
shared. In \f{current}, 67.1\% of distinct guides and inserts appear on one label, 9.2\% appear
on five or more, one appears on 96, and 14.4\% are shared by labels that name more than one
maker. An analysis over labels would count the texts of widely copied medicines many times,
which is why this paper reports over distinct texts.

\begin{figure}[t]
\centering
\includegraphics[width=\textwidth]{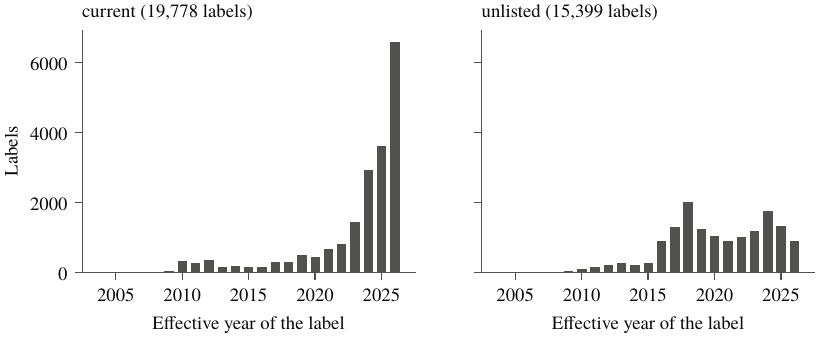}
\caption{Labels by effective year, per configuration.}
\label{fig:years}
\end{figure}

Second, the two configurations differ in age (Figure~\ref{fig:years}). The median effective
year of a \f{current} label is 2025, and two thirds (66.7\%) are dated 2024 or later. The
median for \f{unlisted} labels is 2020. Most \f{unlisted} labels are versions that DailyMed's
release no longer carries, and the source keeps the last version of a label whether or not the
product is still sold. Studies of current patient information should use \f{current}.

% ---------------------------------------------------------------------------------------
\section{Usage notes}
\label{sec:usage}

This section says how to load the corpus and what it can be used for.

\textbf{Loading.} Each table is a configuration on Hugging Face and loads with the
\textit{datasets} library \citep{lhoest2021}, for example
\f{load_dataset("obadakr/plainpairs", "current_labels", split="train")}. A user should choose
one label configuration and one measurement basis and keep to them. Shares and medians should
be taken over distinct texts, through the \f{patient_texts} table, unless the question is
about labels.

\textbf{Health communication research.} The corpus supports questions that earlier samples
were too small to ask: how reading level varies by medicine class, by maker or over time; how
the same risk is worded for the two readers; which headings guides include and omit; and how
patient texts change when a label is revised.

\textbf{Text simplification.} The segment pairs give naturally occurring expert and lay
passages on the same topic, written by the same organization for the same product. They are
topic-level pairs and not sentence alignments. The pairs whose heading passed the hand check
are marked, and the similarity scores let a user set a stricter threshold. A benchmark that
asks a system to rewrite a prescriber section for a general reader, scored against the
official patient segment and a target grade, has been designed and is not yet built.

\textbf{Cross-source and cross-language work.} The cancer summaries allow the same analyses
on a second kind of official text, and their language pairs join English and Spanish versions
of the same summary.

\textbf{What not to do.} The corpus must not be used as a source of information about
medicines. Labels in it may be out of date or superseded. Reading levels from the two bases
must not be mixed, and figures for the two label configurations must not be pooled.

% ---------------------------------------------------------------------------------------
\section{Limitations}
\label{sec:limits}

This section lists what limits the claims in this paper, starting with the largest.

\textbf{Formulas.} Flesch-Kincaid and SMOG see word and sentence length and nothing else.
They do not see layout, type size, the order of information or whether a reader can act on the
text, all of which matter for patient materials \citep{redish2000,shoemaker2014}. Syllables are
counted by rule. The grades are comparable within the corpus and should not be compared with
grades computed by other software on other preparations of the text, which is itself one of
the results above.

\textbf{Structure.} The XML keeps more layout than the bulk files, but headings are found by a
rule about bold and underlined lines, and a heading set in plain type is read as a sentence.
Structure is available for \f{current} labels only.

\textbf{Topic-level alignment.} A segment is matched to whole sections. The corpus does not
say which sentences of a section a patient sentence corresponds to, and three of the ten
heading types have no reliable mapping.

\textbf{Rights.} The copyright marks are found by matching three patterns. A label can assert
rights in other words, and a mark does not establish that reuse is restricted. The switch is a
cautious default and not a legal finding.

% ---------------------------------------------------------------------------------------
\section{Conclusion}
\label{sec:conclusion}

PlainPairs puts the two official texts of U.S.\ prescription medicines side by side: 35,177
labels, 21,416 distinct patient texts, and 250,829 segments matched to prescriber sections,
with measures and wording counts on both sides and a second pair from cancer summaries. The
first release shows that Medication Guides sit near grade 9 on one common formula and near
grade 11 on another, that the flattened text of the bulk files overstates both by one to two
grades, and that patient texts address the reader directly while leaving out nearly all the
numbers the prescriber is given. I expect the corpus to be useful for work on how medicine
information is worded for patients, on automatic rewriting of medical text, and, if the
proposed one-page patient document becomes a rule, on how patient labeling changes.

\section*{Data and code availability}

The corpus is at \url{https://huggingface.co/datasets/obadakr/plainpairs} (version 0.1.0, tag
\f{v0.1.0}), with a mirror at \url{https://www.kaggle.com/datasets/obadakraishan/plainpairs}.
The software, tests and documents are at \url{https://github.com/obadaKraishan/plainpairs} and
archived at \url{https://doi.org/10.5281/zenodo.24267450}. The project's own additions
(measures, counts, flags, alignments and the layout of the tables) are dedicated to the public
domain under CC0 1.0, and the software is under the Apache License 2.0. The label text comes
from openFDA, and the structure from DailyMed, courtesy of the U.S.\ National Library of
Medicine. The cancer summaries were originally published by the National Cancer Institute.
None of these organizations made, endorsed or checked the corpus. The scripts that computed
the derived figures and drew the figures of this paper from the public tables are included
with its source.

\bibliographystyle{abbrvnat}
\bibliography{refs}

@article{flesch1948,
  author  = {Flesch, Rudolph},
  title   = {A new readability yardstick},
  journal = {Journal of Applied Psychology},
  year    = {1948},
  volume  = {32},
  number  = {3},
  pages   = {221--233},
  doi     = {10.1037/h0057532}
}

@techreport{kincaid1975,
  author      = {Kincaid, J. Peter and Fishburne, Jr., Robert P. and Rogers, Richard L. and Chissom, Brad S.},
  title       = {Derivation of New Readability Formulas ({Automated Readability Index}, {Fog Count} and {Flesch Reading Ease Formula}) for {Navy} Enlisted Personnel},
  institution = {Naval Technical Training Command, Research Branch},
  address     = {Millington, TN},
  year        = {1975},
  type        = {Research Branch Report},
  number      = {8-75},
  doi         = {10.21236/ADA006655}
}

@article{mclaughlin1969,
  author  = {McLaughlin, G. Harry},
  title   = {{SMOG} grading: A new readability formula},
  journal = {Journal of Reading},
  year    = {1969},
  volume  = {12},
  number  = {8},
  pages   = {639--646}
}

@article{wolf2006,
  author  = {Wolf, Michael S. and Davis, Terry C. and Shrank, William H. and Neuberger, Marolee and Parker, Ruth M.},
  title   = {A critical review of {FDA}-approved {Medication Guides}},
  journal = {Patient Education and Counseling},
  year    = {2006},
  volume  = {62},
  number  = {3},
  pages   = {316--322},
  doi     = {10.1016/j.pec.2006.06.010}
}

@article{wolf2012,
  author  = {Wolf, Michael S. and King, Jennifer and Wilson, Elizabeth A. H. and Curtis, Laura M. and Bailey, Stacy Cooper and Duhig, James and Russell, Allison and Bergeron, Ashley and Daly, Amanda and Parker, Ruth M. and Davis, Terry C. and Shrank, William H. and Lambert, Bruce},
  title   = {Usability of {FDA}-approved {Medication Guides}},
  journal = {Journal of General Internal Medicine},
  year    = {2012},
  volume  = {27},
  number  = {12},
  pages   = {1714--1720},
  doi     = {10.1007/s11606-012-2068-7}
}

@article{jones2021,
  author  = {Jones, Paul R. and Walther, Elisabeth J. and Nguyen, Andrew and Santiago, Jonas and Pearsall, Bryon M.},
  title   = {Biennial analysis of {Medication Guide} length and estimated readability for new molecular entity drugs, 2011--2017},
  journal = {Therapeutic Innovation \& Regulatory Science},
  year    = {2021},
  volume  = {55},
  number  = {5},
  pages   = {918--925},
  doi     = {10.1007/s43441-021-00270-3}
}

@article{meyer2023,
  author  = {Meyer, Claudia and Adkins, Daniel and Pal, Koyena and Galici, Ruggero and Garcia-Agundez, Augusto and Eickhoff, Carsten},
  title   = {Neural text generation in regulatory medical writing},
  journal = {Frontiers in Pharmacology},
  year    = {2023},
  volume  = {14},
  pages   = {1086913},
  doi     = {10.3389/fphar.2023.1086913}
}

@article{attal2023,
  author  = {Attal, Kush and Ondov, Brian and Demner-Fushman, Dina},
  title   = {A dataset for plain language adaptation of biomedical abstracts},
  journal = {Scientific Data},
  year    = {2023},
  volume  = {10},
  number  = {1},
  pages   = {8},
  doi     = {10.1038/s41597-022-01920-3}
}

@inproceedings{basu2023,
  author    = {Basu, Chandrayee and Vasu, Rosni and Yasunaga, Michihiro and Yang, Qian},
  title     = {{Med-EASi}: Finely annotated dataset and models for controllable simplification of medical texts},
  booktitle = {Proceedings of the AAAI Conference on Artificial Intelligence},
  year      = {2023},
  volume    = {37},
  pages     = {14093--14101},
  doi       = {10.1609/aaai.v37i12.26649}
}

@inproceedings{devaraj2021,
  author    = {Devaraj, Ashwin and Marshall, Iain and Wallace, Byron and Li, Junyi Jessy},
  title     = {Paragraph-level simplification of medical texts},
  booktitle = {Proceedings of the 2021 Conference of the North American Chapter of the Association for Computational Linguistics: Human Language Technologies},
  year      = {2021},
  pages     = {4972--4984},
  doi       = {10.18653/v1/2021.naacl-main.395}
}

@inproceedings{cao2020,
  author    = {Cao, Yixin and Shui, Ruihao and Pan, Liangming and Kan, Min-Yen and Liu, Zhiyuan and Chua, Tat-Seng},
  title     = {Expertise style transfer: A new task towards better communication between experts and laymen},
  booktitle = {Proceedings of the 58th Annual Meeting of the Association for Computational Linguistics},
  year      = {2020},
  pages     = {1061--1071},
  doi       = {10.18653/v1/2020.acl-main.100}
}

@article{guo2024,
  author  = {Guo, Yue and Qiu, Wei and Leroy, Gondy and Wang, Sheng and Cohen, Trevor},
  title   = {Retrieval augmentation of large language models for lay language generation},
  journal = {Journal of Biomedical Informatics},
  year    = {2024},
  volume  = {149},
  pages   = {104580},
  doi     = {10.1016/j.jbi.2023.104580}
}

@inproceedings{goldsack2022,
  author    = {Goldsack, Tomas and Zhang, Zhihao and Lin, Chenghua and Scarton, Carolina},
  title     = {Making science simple: Corpora for the lay summarisation of scientific literature},
  booktitle = {Proceedings of the 2022 Conference on Empirical Methods in Natural Language Processing},
  year      = {2022},
  pages     = {10589--10604},
  doi       = {10.18653/v1/2022.emnlp-main.724}
}

@article{campillos2022,
  author  = {Campillos-Llanos, Leonardo and Terroba Reinares, Ana R. and Zakhir Puig, Sof{\'i}a and Valverde-Mateos, Ana and Capllonch-Carri{\'o}n, Adri{\'a}n},
  title   = {Building a comparable corpus and a benchmark for {Spanish} medical text simplification},
  journal = {Procesamiento del Lenguaje Natural},
  year    = {2022},
  volume  = {69},
  pages   = {189--196}
}

@article{kasshout2016,
  author  = {Kass-Hout, Taha A. and Xu, Zhiheng and Mohebbi, Matthew and Nelsen, Hans and Baker, Adam and Levine, Jonathan and Johanson, Elaine and Bright, Roselie A.},
  title   = {{OpenFDA}: An innovative platform providing access to a wealth of {FDA}'s publicly available data},
  journal = {Journal of the American Medical Informatics Association},
  year    = {2016},
  volume  = {23},
  number  = {3},
  pages   = {596--600},
  doi     = {10.1093/jamia/ocv153}
}

@article{shoemaker2014,
  author  = {Shoemaker, Sarah J. and Wolf, Michael S. and Brach, Cindy},
  title   = {Development of the {Patient Education Materials Assessment Tool} ({PEMAT}): A new measure of understandability and actionability for print and audiovisual patient information},
  journal = {Patient Education and Counseling},
  year    = {2014},
  volume  = {96},
  number  = {3},
  pages   = {395--403},
  doi     = {10.1016/j.pec.2014.05.027}
}

@article{wang2013,
  author  = {Wang, Lih-Wern and Miller, Michael J. and Schmitt, Michael R. and Wen, Frances K.},
  title   = {Assessing readability formula differences with written health information materials: Application, results, and recommendations},
  journal = {Research in Social and Administrative Pharmacy},
  year    = {2013},
  volume  = {9},
  number  = {5},
  pages   = {503--516},
  doi     = {10.1016/j.sapharm.2012.05.009}
}

@article{barrio2008,
  author  = {Barrio-Cantalejo, I. M. and Sim{\'o}n-Lorda, P. and Melguizo, M. and Escalona, I. and Mariju{\'a}n, M. I. and Hernando, P.},
  title   = {Validaci{\'o}n de la {Escala INFLESZ} para evaluar la legibilidad de los textos dirigidos a pacientes},
  journal = {Anales del Sistema Sanitario de Navarra},
  year    = {2008},
  volume  = {31},
  number  = {2},
  pages   = {135--152},
  doi     = {10.4321/s1137-66272008000300004}
}

@article{gebru2021,
  author  = {Gebru, Timnit and Morgenstern, Jamie and Vecchione, Briana and Vaughan, Jennifer Wortman and Wallach, Hanna and Daum{\'e} III, Hal and Crawford, Kate},
  title   = {Datasheets for datasets},
  journal = {Communications of the ACM},
  year    = {2021},
  volume  = {64},
  number  = {12},
  pages   = {86--92},
  doi     = {10.1145/3458723}
}

@article{wilson1927,
  author  = {Wilson, Edwin B.},
  title   = {Probable inference, the law of succession, and statistical inference},
  journal = {Journal of the American Statistical Association},
  year    = {1927},
  volume  = {22},
  number  = {158},
  pages   = {209--212},
  doi     = {10.1080/01621459.1927.10502953}
}

@article{berkman2011,
  author  = {Berkman, Nancy D. and Sheridan, Stacey L. and Donahue, Katrina E. and Halpern, David J. and Crotty, Karen},
  title   = {Low health literacy and health outcomes: An updated systematic review},
  journal = {Annals of Internal Medicine},
  year    = {2011},
  volume  = {155},
  number  = {2},
  pages   = {97--107},
  doi     = {10.7326/0003-4819-155-2-201107190-00005}
}

@article{davis2006,
  author  = {Davis, Terry C. and Wolf, Michael S. and Bass, Pat F. and Thompson, Jason A. and Tilson, Hugh H. and Neuberger, Marolee and Parker, Ruth M.},
  title   = {Literacy and misunderstanding prescription drug labels},
  journal = {Annals of Internal Medicine},
  year    = {2006},
  volume  = {145},
  number  = {12},
  pages   = {887--894},
  doi     = {10.7326/0003-4819-145-12-200612190-00144}
}

@article{shrank2007,
  author  = {Shrank, William H. and Avorn, Jerry},
  title   = {Educating patients about their medications: The potential and limitations of written drug information},
  journal = {Health Affairs},
  year    = {2007},
  volume  = {26},
  number  = {3},
  pages   = {731--740},
  doi     = {10.1377/hlthaff.26.3.731}
}

@article{redish2000,
  author  = {Redish, Janice},
  title   = {Readability formulas have even more limitations than {Klare} discusses},
  journal = {ACM Journal of Computer Documentation},
  year    = {2000},
  volume  = {24},
  number  = {3},
  pages   = {132--137},
  doi     = {10.1145/344599.344637}
}

@article{xu2015,
  author  = {Xu, Wei and Callison-Burch, Chris and Napoles, Courtney},
  title   = {Problems in current text simplification research: New data can help},
  journal = {Transactions of the Association for Computational Linguistics},
  year    = {2015},
  volume  = {3},
  pages   = {283--297},
  doi     = {10.1162/tacl_a_00139}
}

@article{althanyyan2021,
  author  = {Al-Thanyyan, Suha S. and Azmi, Aqil M.},
  title   = {Automated text simplification: A survey},
  journal = {ACM Computing Surveys},
  year    = {2021},
  volume  = {54},
  number  = {2},
  pages   = {1--36},
  doi     = {10.1145/3442695}
}

@article{salton1988,
  author  = {Salton, Gerard and Buckley, Christopher},
  title   = {Term-weighting approaches in automatic text retrieval},
  journal = {Information Processing \& Management},
  year    = {1988},
  volume  = {24},
  number  = {5},
  pages   = {513--523},
  doi     = {10.1016/0306-4573(88)90021-0}
}

@article{dice1945,
  author  = {Dice, Lee R.},
  title   = {Measures of the amount of ecologic association between species},
  journal = {Ecology},
  year    = {1945},
  volume  = {26},
  number  = {3},
  pages   = {297--302},
  doi     = {10.2307/1932409}
}

@article{candelario2017,
  author  = {Candelario, Danielle M. and Vazquez, Victoria and Jackson, William and Reilly, Timothy},
  title   = {Completeness, accuracy, and readability of {Wikipedia} as a reference for patient medication information},
  journal = {Journal of the American Pharmacists Association},
  year    = {2017},
  volume  = {57},
  number  = {2},
  pages   = {197--200.e1},
  doi     = {10.1016/j.japh.2016.12.063}
}

@article{moore2023,
  author  = {Moore, John E. and Tang, Ka Wah Kelly and Millar, Beverley C.},
  title   = {Improving health literacy of antifungal use: Comparison of the readability of antifungal medicines information from {Australia}, {EU}, {UK}, and {US}},
  journal = {Medical Mycology},
  year    = {2023},
  volume  = {61},
  number  = {9},
  pages   = {myad084},
  doi     = {10.1093/mmy/myad084}
}

@article{pires2015,
  author  = {Pires, Carla and Vig{\'a}rio, Marina and Cavaco, Afonso},
  title   = {Readability of medicinal package leaflets: A systematic review},
  journal = {Revista de Sa{\'u}de P{\'u}blica},
  year    = {2015},
  volume  = {49},
  doi     = {10.1590/s0034-8910.2015049005559}
}

@article{wilkinson2016,
  author  = {Wilkinson, Mark D. and Dumontier, Michel and Aalbersberg, IJsbrand Jan and Appleton, Gabrielle and Axton, Myles and others},
  title   = {The {FAIR Guiding Principles} for scientific data management and stewardship},
  journal = {Scientific Data},
  year    = {2016},
  volume  = {3},
  pages   = {160018},
  doi     = {10.1038/sdata.2016.18}
}

@inproceedings{lhoest2021,
  author    = {Lhoest, Quentin and Villanova del Moral, Albert and Jernite, Yacine and Thakur, Abhishek and von Platen, Patrick and Patil, Suraj and Chaumond, Julien and Drame, Mariama and Plu, Julien and Tunstall, Lewis and others},
  title     = {Datasets: A community library for natural language processing},
  booktitle = {Proceedings of the 2021 Conference on Empirical Methods in Natural Language Processing: System Demonstrations},
  year      = {2021},
  pages     = {175--184},
  doi       = {10.18653/v1/2021.emnlp-demo.21}
}

@article{sridharan2024,
  author  = {Sridharan, Kannan and Sivaramakrishnan, Gowri},
  title   = {Enhancing readability of {USFDA} patient communications through large language models: A proof-of-concept study},
  journal = {Expert Review of Clinical Pharmacology},
  year    = {2024},
  volume  = {17},
  number  = {8},
  pages   = {731--741},
  doi     = {10.1080/17512433.2024.2363840}
}

@article{chan2018,
  author  = {Chan, Hilda W. and Russell, Andrea M. and Smith, Meredith Y.},
  title   = {What is the quality of drug safety information for patients: An analysis of {REMS} educational materials},
  journal = {Pharmacoepidemiology and Drug Safety},
  year    = {2018},
  volume  = {27},
  number  = {9},
  pages   = {969--978},
  doi     = {10.1002/pds.4614}
}

@article{noble2022,
  author  = {Noble, Adam J. and Haddad, Sara and Coleman, Niamh and Marson, Anthony G.},
  title   = {Worth the paper they are printed on? {Findings} from an independent evaluation of the understandability of patient information leaflets for antiseizure medications},
  journal = {Epilepsia},
  year    = {2022},
  volume  = {63},
  number  = {8},
  pages   = {2130--2143},
  doi     = {10.1111/epi.17299}
}

@misc{cfr208,
  year         = {2026},
  author       = {{U.S. Food and Drug Administration}},
  title        = {Medication Guides for Prescription Drug Products. {Code of Federal Regulations}, Title 21, Part 208},
  howpublished = {\url{https://www.ecfr.gov/current/title-21/part-208}},
  note         = {Read 15 November 2026}
}

@misc{pmi2023,
  author       = {{U.S. Food and Drug Administration}},
  title        = {Medication Guides: {Patient Medication Information}. {Proposed} rule},
  howpublished = {Federal Register 88(104):35694, 31 May 2023. Docket No. FDA-2019-N-5959},
  year         = {2023},
  note         = {\url{https://www.govinfo.gov/content/pkg/FR-2023-05-31/html/2023-11354.htm}}
}

@misc{openfda_terms,
  year         = {2026},
  author       = {{openFDA}},
  title        = {Terms of service, and drug product labeling bulk downloads},
  howpublished = {\url{https://open.fda.gov/terms/}},
  note         = {Read 15 November 2026}
}

@misc{dailymed,
  year         = {2026},
  author       = {{U.S. National Library of Medicine}},
  title        = {{DailyMed}: {SPL} resources, download all drug labels},
  howpublished = {\url{https://dailymed.nlm.nih.gov/dailymed/spl-resources-all-drug-labels.cfm}},
  note         = {Read 15 November 2026}
}

@misc{nci_reuse,
  year         = {2026},
  author       = {{National Cancer Institute}},
  title        = {Reuse of {NCI} information},
  howpublished = {\url{https://www.cancer.gov/policies/copyright-reuse}},
  note         = {Read 15 November 2026}
}

@misc{plainpairs_data,
  author       = {Kraishan, Obada},
  title        = {{PlainPairs}: A research corpus pairing the prescriber and patient texts of {US} medicine labels (Version 0.1.0) [Data set]},
  howpublished = {Hugging Face, \url{https://huggingface.co/datasets/obadakr/plainpairs}},
  year         = {2026}
}

@misc{plainpairs_code,
  author       = {Kraishan, Obada},
  title        = {{PlainPairs}: Software and documents for a research corpus pairing the prescriber and patient texts of {US} medicine labels (v0.1.0)},
  howpublished = {Zenodo},
  year         = {2026},
  doi          = {10.5281/zenodo.24267450}
}

\appendix

\section{Main fields}
\label{app:fields}

Table~\ref{tab:fields} lists the main fields of the record tables. The full schema is in the
repository.

\begin{table}[H]
\centering\small
\caption{Main fields of the record tables.}
\label{tab:fields}
\begin{tabularx}{\textwidth}{@{}L{2.5cm}L{5.4cm}Y@{}}
\toprule
Table & Field & Meaning \\
\midrule
\f{labels} & \f{set_id}, \f{version}, \f{effective_date} & the label, its version and its date \\
 & \f{label_format} & newer or older prescription format \\
 & \f{medicine_key}, \f{medicine_key_source} & generic names, and whether they are the label's own or borrowed through a shared patient text \\
 & \f{listed_in_ndc_directory}, \f{served_by_dailymed} & status on the two official lists \\
 & \f{configuration}, \f{unlisted_reason} & \f{current} or \f{unlisted}, and why \\
 & \f{has_copyright_mark} & a copyright mark in any text field \\
 & \f{p_<section>}, \f{p_<section>_structured} & measures of each prescriber section, on the flattened text and on structure \\
 & \f{t_<text>}, \f{t_<text>_structured}, \f{t_<text>_id} & measures of each patient text, and its row in \f{patient_texts} \\
\f{patient_texts} & \f{patient_text_id}, \f{kind} & the distinct text and its kind \\
 & \f{n_labels}, \f{crosses_makers} & labels that share it; whether they name more than one maker \\
 & \f{headings_found} & standard headings present \\
 & \f{measures}, \f{measures_structured} & measures on the two bases \\
\f{section_pairs} & \f{heading_key}, \f{segment_index} & the heading and the position of the segment \\
 & \f{mapped_sections}, \f{mapped_present} & sections the heading maps to; those the label has \\
 & \f{best_section}, \f{best_similarity} & the most similar section and its score \\
 & \f{mapped_best_similarity}, \f{best_is_mapped} & the best score among mapped sections; whether the two agree \\
 & \f{rule_status} & whether the heading's mapping passed the hand check \\
\f{codes} & \f{section}, \f{side} & the text, and whether it is a patient or a prescriber text \\
 & \f{you_words}, \f{sentences_with_you} & direct address \\
 & \f{command_sentences}, \f{distinct_actions} & instructions as commands \\
 & \f{percentages}, \f{one_in_n}, \f{frequencies}, \f{doses} & numbers by kind \\
 & \f{contact_sentences} & sentences that say whom to contact \\
\f{nci_summaries} & \f{url}, \f{language}, \f{version}, \f{nci_updated} & the page, its language and version, and the institute's update date \\
 & \f{credit_line} & the credit to print with any reuse \\
 & \f{left_out} & counts of the parts left out \\
\bottomrule
\end{tabularx}
\end{table}

\section{Reading levels of prescriber sections}
\label{app:sections}

Table~\ref{tab:sections} gives the figures behind Figure~\ref{fig:sections}.

\begin{table}[H]
\centering\small
\caption{Prescriber sections in \f{current}, over distinct texts, measured on structure.}
\label{tab:sections}
\begin{tabular}{@{}lrrrr@{}}
\toprule
Section & Distinct texts & Median words & Flesch-Kincaid (middle half) & SMOG \\
\midrule
Boxed warning & 5,326 & 215 & 14.59 (13.02 to 16.11) & 15.88 \\
Indications and usage & 9,162 & 136 & 15.82 (13.87 to 17.87) & 16.66 \\
Dosage and administration & 10,687 & 523 & 12.65 (11.21 to 14.28) & 14.33 \\
Contraindications & 7,915 & 80 & 16.18 (14.07 to 18.16) & 15.25 \\
Warnings and precautions & 11,557 & 1,686 & 16.12 (15.18 to 16.91) & 17.01 \\
Adverse reactions & 11,689 & 841 & 15.53 (14.33 to 16.76) & 16.19 \\
Drug interactions & 8,661 & 243 & 16.59 (15.37 to 17.86) & 17.12 \\
Use in specific populations & 8,528 & 1,027 & 15.61 (14.81 to 16.35) & 16.65 \\
Patient counseling information & 10,990 & 426 & 13.79 (11.97 to 15.16) & 14.96 \\
Description & 10,716 & 148 & 15.09 (13.77 to 16.39) & 16.08 \\
How supplied and storage & 18,543 & 87 & 6.58 (5.01 to 8.33) & 9.57 \\
\bottomrule
\end{tabular}
\end{table}

\end{document}